\documentclass[conference, a4paper]{IEEEtran}
\IEEEoverridecommandlockouts

\usepackage{cite}
\usepackage{amsmath,amssymb,amsfonts}
\usepackage{algorithmic}
\usepackage{graphicx}
\usepackage{textcomp}
\usepackage{xcolor}
\def\BibTeX{{\rm B\kern-.05em{\sc i\kern-.025em b}\kern-.08em
    T\kern-.1667em\lower.7ex\hbox{E}\kern-.125emX}}

\usepackage{times}
\usepackage{latexsym}
\usepackage{graphicx}
\usepackage{caption}
\usepackage{subcaption}
\usepackage{algorithm}
\usepackage{algorithmic}
\usepackage{amsmath}
\DeclareMathOperator*{\argmax}{arg\,max}

\usepackage{multirow}
\usepackage{booktabs}
\usepackage{tabularx}
\usepackage{enumitem}  
\usepackage{calc}  

\usepackage{balance}

\newcommand\clearrow{\global\let\rowmac\relax}
\clearrow

\begin{document}

\title{A Bayesian Approach to Reinforcement Learning of Vision-Based Vehicular Control\\
\thanks{Funded by the Wallenberg AI, Autonomous Systems and Software Program
	(WASP).}
}

\author{\IEEEauthorblockN{Zahra Gharaee$^{*}$$^{1}$}
\and
\IEEEauthorblockN{Karl Holmquist$^{1}$}

\and
\IEEEauthorblockN{Linbo He$^{1}$}

\and
\IEEEauthorblockN{Michael Felsberg$^{1}$}

\thanks{$^{*}$ Corresponding Author}%

\thanks{$^{1}$Computer Vision Laboratory (CVL), 
	Department of Electrical Engineering, University of Link\"oping, 
	58183  Link\"oping, Sweden. 
	Emails:
	{\tt\small \{zahra.gharaee, karl.holmquist, linbo.he, michael.felsberg\}@liu.se}}
}

\maketitle

\begin{abstract}
In this paper, we present a state-of-the-art reinforcement learning method for autonomous driving. Our approach employs temporal difference learning in a Bayesian framework to learn vehicle control signals from sensor data. The agent has access to images from a forward facing camera, which are pre-processed to generate semantic segmentation maps. We trained our system using both ground truth and estimated semantic segmentation input.

Based on our observations from a large set of experiments, we conclude that training the system on ground truth input data leads to better performance than  training the system on estimated input even if estimated input is used for evaluation.

The system is trained and evaluated in a realistic simulated urban environment using the CARLA simulator. The simulator also contains a benchmark that allows for comparing to other systems and methods. The required training time of the system is shown to be lower and the performance on the benchmark superior to competing approaches.

\end{abstract}


\begin{figure*}[h]
	\centering
	\begin{subfigure}{\linewidth}
		\includegraphics[width=\linewidth]{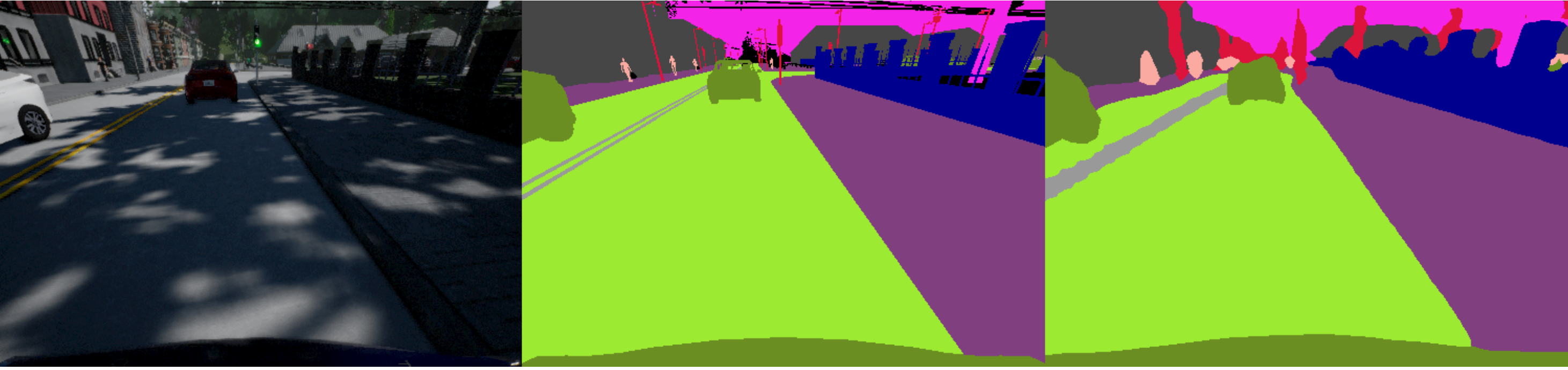}
	\end{subfigure}
	\caption{Example images from the CARLA simulator, RGB-image (left), ground-truth semantic segmentation from simulator (center) and estimated semantic segmentation from EncNet \cite{HangZhang} (right). Top row shows driving during raining and the bottom row shows the sunny noon weather condition.}
	\label{fig:Semantics}
\end{figure*}

\section{Introduction}

Autonomous driving is a complex problem with enormous application value. In this paper we address the essential functionality of road following by a reinforcement learning (RL) based approach.

Systems learning in an end-to-end manner have shown their ability to accurately predict steering signals of the vehicle \cite{chen2017end}. However, in particular deep networks inherently require a large amount of training data. This problem can be somewhat mitigated by the use of simulators, which also allows safe exploration of policies without the risk of harming the agent or other entities in the environment. Furthermore, the increasingly more realistic simulators facilitates moving from the virtual to the real environments. 

We use the CARLA simulator \cite{Dosovitskiy17} for training and evaluating our system in a realistic driving environment. The system is trained using two different settings for the inputs, (a) ground-truth and (b) estimated semantic segmentation to generate two sets of models. The estimated semantic segmentation is generated by a pre-trained Context Encoding Network (EncNet \cite{HangZhang}). We evaluate our two model sets using both ground-truth and estimated input, creating a total of four evaluation results reaching state-of-the-art performance. We show that keeping the preprocessing frozen facilitates efficient training of our agent in online mode. However, the most interesting result is that the trained system with ground-truth input performs superior to the trained system with estimated input data.

The main contributions of this paper are:\\
(1) We propose a method for learning realistic autonomous driving in the CARLA simulator based on a Bayesian approach to Reinforcement Learning \cite{firouzi2012interactive}. Our proposed method splits learning into offline preprocessing of the raw data and online reinforcement learning of the control.\\
(2) We implement an algorithm based on this method that reaches state-of-the-art performance, which shows that the proposed method can learn the driving task efficiently.\\
(3) We suggest training the system using ground-truth data available from the CARLA simulator since the test performance is better than when trained on estimated data.\\

\section{Related Work}

The problem of autonomous driving constitutes of multiple levels, from the top-level route planning \cite{mirowski2018learning} to the low-level control. In this paper we focus on learning the mapping from perception to control signals.

There has been a large success in complex tasks, from games \cite{mnih2013playing,silver2016mastering} to continuous control problems \cite{brockman2016openai,gu2017deep}, with the introduction of model-free deep RL. 
Q-learning approaches are popular for discrete action spaces either in their tabular form or in their deep formulation using neural networks as function approximators \cite{mnih2013playing, sutton1998introduction}.
On the other hand, the majority of deep RL approaches for continuous control problems are based on the actor-critic family \cite{mnih2016asynchronous, haarnoja2018soft}.

Using neural networks for function approximation faces two main challenges. First, these methods are expensive in terms of their sample complexity for data collection in high dimensional input space. 
Second, they are sensitive to the model hyper-parameters such as learning rates and exploration constants.

Since deep RL methods usually require excessive data collection through interaction with the environment, they can be prohibitively expensive in real-world scenarios and also constraining the acceptable exploratory behavior. As such, there has been a large increase in available realistic simulators, such as CARLA \cite{Dosovitskiy17}. These often provide simulated raw measurements from different sensors, e.g. RGB-cameras, semantic information, LiDARs, positioning and dense-depth measures.

However, a major problem when using simulated data is the domain shift when moving from the virtual to the real world. This problem can be addressed by adapting the simulated data to a format closer to the real data \cite{BMVC2017_11} or generating simulated data from real \cite{yang2018real}. A commonly applied approach is to map the input space to a semantic representation \cite{muller2018driving}. However, while this work suggests to only segment the world into road and non-road, we additionally introduce obstacles and road line in our semantic representation. Another difference is that we directly predict the control signals rather than way-points used for navigation. A comparison to \cite{muller2018driving} is not feasible since the code is not publicly available.

To address the complexity and uncertainty in a realistic environment, probabilistic methods have been shown to be useful for many types of problems \cite{shi2019soft}. Recent probabilistic approaches to reinforcement learning have been formulated in terms of maximum entropy RL. These methods add an entropy cost to the predicted policy to avoid fast convergence to a single action and encouraging more robust and exploratory policies during training \cite{o2016combining}. However, these approaches become highly unstable if the relative scaling between reward and entropy is incorrectly set \cite{haarnoja2018entropy}.

Instead, we based our method on a Bayesian approach to reinforcement learning \cite{firouzi2012interactive}. In this approach, we use an online reinforcement-based clustering of the agent's perceptual space by a Gaussian mixture model (GMM). These mixture components are matched with a Bayesian formulation to benefit the approach in terms of generalization and adaptability. Rather than adapting to a simulated robotic platform \cite{gharaee2016attention}, we test our method in a realistic driving scenario using the CARLA simulator. We use a learned preprocessing module based on the semantic segmentation to facilitate efficient online training of the method. Furthermore, we study the impact of the noise introduced by the estimation errors in the preprocessing module for both training and the deployment of the trained system.

Besides other RL-based algorithms, we also compare our method to a supervised Imitation Learning approach \cite{codevilla2019exploring}. The imitation learning approaches are however limited to that expert demonstrations exist for the relevant scenarios. This is problematic in dangerous and rare situations, which are often avoided before occurring. Therefore, it is difficult to rely solely on the expert demonstrations.

\section{Method}

The problem of driving in a city environment is a complex task composed of a prediction problem (estimating the state-action value function for a given policy) as well as a control problem (finding the optimal policy). Therefore, the target value to be estimated for the agent is the probability of a specific action given the input state. These probabilities are used for decision making. 

\subsection{Bayesian approach to Reinforcement Learning}

In this section we present the Bayesian approach to Reinforcement Learning introduced by \cite{firouzi2012interactive}. This framework integrates a Gaussian Mixture Model (GMM) with a reinforcement learning approach. We assume that at each time step $t$, the agent perceives the world through the state $s_t$ and models the action-conditioned state probability by a GMM with a set of mixture components $\mathcal{M}$. Based on the perceived state, the agent makes a decision and performs an action $a_t$ from the action set $\mathcal{A}$. The agent's decision is evaluated by a reward signal $r_t$, which is used to train the system. 

We assume stochastic variables $a \in\mathcal{A}$ and $m \in\mathcal{M}$ to calculate the conditional probability $p(a|s_t)$. These probabilities are used to select the best decision at a given state and are estimated as:

\begin{equation}
p(a|s_t) \propto  p(a) \sum_{m\in\mathcal{M}} p(s_t|m) p(m|a).
\label{eq:pcx}
\end{equation}

Due to the unknown parameters of the GMM we are using the multi-variate t-distribution as proposed by \cite{firouzi2012interactive} to estimate $p(s_t|m)$, the likelihood of the state $s_t$ given the mixture component $m$. 

Next term to be estimated in (\ref{eq:pcx}) is $p(m|a)$, the probability of mixture component $m$ given the action $a$. As suggested by \cite{firouzi2012interactive}, this probability is parameterized by a state-action value function $Q = [ q_{m,a} ]: (|\mathcal{M}| \times |\mathcal{A}| )$ ($|\cdot|$ represents the cardinality of the respective set), and learned by reinforcement learning. To calculate the probabilities $p(m|a)$ and $p(a)$, the elements of $Q$ have to be non-negative, thus \cite{firouzi2012interactive} proposed to use the offset:

\begin{equation}
\hat{q} = \frac{|\min \{Q\}|}{1 + |\min \{Q\}|} - \min \{Q\}.
\label{eq:qmat}
\end{equation}

The expressions for the probabilities $p(m|a)$ and $p(a)$ read as follows:

\begin{equation}
p(m|a) \propto (q_{m,a} + \hat{q}).
\label{eq:pma}
\end{equation}

\begin{equation}
p(a) \propto \sum_{m\in\mathcal{M}}(q_{m,a}+\hat{q}).
\label{eq:pa}
\end{equation}

To train the system, a temporal difference learning (TD) approach is used with the loss:

\begin{equation}
TD_{\textrm{error}} = r_t +  \gamma Q(a_{t+1}, m_{t+1}) - Q(a_{t}, m_{t}),
\label{tderr}
\end{equation}

where $r_t$ is the reward signal and $\gamma$ is the forgetting factor. Term $a_{t}$ shows the action performed at time step $t$ and $m_{t}$ is the most similar component to the state $s_t$ given by $\ell_\infty $ norm between the state $s_t$ and the mean vectors of the existing mixture components. Term $a_{t+1}$ shows the most probable action in the next time step $t+1$:

\begin{equation}
a_{t+1} =  \argmax_a p(a|s_{t+1}).
\label{at}
\end{equation}

The most likely component is $m_{t+1}$, based on the probability:

\begin{equation}
p(m|a_{t+1}, s_{t+1}) \propto
p(s_{t+1}|m)p(m|a_{t+1}).
\label{eq:pmas}
\end{equation}

Finally the system parameters are learned by Q-Learning using (\ref{eq:Qupdate}), where $\alpha$ corresponds to a decaying learning rate and $w$ allows for a soft updates despite the greedy choice of component $m_{t+1}$:

\begin{equation}
Q(m_t, a_t) \gets Q(m_t, a_t) + \alpha w TD_{\textrm{error}}.
\label{eq:Qupdate}
\end{equation}

To calculate $w$, the $TD_{\textrm{error}}$ is evaluated using two boundaries: a lower threshold, $T_{l}$, showing a bad decision and an upper threshold, $T_{u}$, to show a good one:

\begin{equation}
w = 
\begin{cases}
p(m_t | a_{t}, s_{t}), 		& \text{if } TD_{\textrm{error}} > T_{u} \\ 
p(m_t|\neg a_{t}, s_{t}),	& \text{else if } TD_{\textrm{error}} < T_{l} \\
p(m_t | s_{t}),			& \text{else,}
\end{cases}
\label{eq:weight}
\end{equation}
where $p(m | \neg a, s_{t})$ is the probability for component $m$ given not performing action $a$ at state $s_{t}$:

\begin{equation}
p(m|\neg a, s_{t}) \propto p(s_{t}|m)(1-p(m|a)).
\label{pmnas}
\end{equation}

To update the parameters, two criteria are used: 
\begin{enumerate}
	\item Similarity measure $d_t$ given by the $\ell_\infty $ norm between the state $s_t$ and the mean vectors of $m$.
	\item Evaluation criterion given by $TD_{\textrm{error}} < T_{l}$, to determine if the action was a bad choice. 
	
\end{enumerate}
If the observed state is not similar to any of the existing components $m$ and the action $a$ is a bad choice, then a new component centered at $s_{t+1}$ is created otherwise the parameters of $m_t$ are updated, see Algorithm \ref{alg:system}.

\subsection{Reinforcement Learning method for vehicular control}
\label{BRL}

In this section we propose a method that applies the RL approach introduced in Section \ref{BRL} to learn a realistic driving task using the CARLA simulator. This includes design and implementation of the input and the reward signals as well as a policy for decision making. The system is constituted by one agent, which receives a continuous state-vector as the input and predicts discrete actions. The overview of our RL method is described by Algorithm. \ref{alg:system}.

\begin{algorithm}[tbh]
	\caption{Overview of RL method according to \cite{firouzi2012interactive}}
	\label{alg:system}
	\textbf{Input}: Initialized model.\\
	\textbf{Output}: Trained model.\\
	\textbf{t$_\mathbf{max}$}: Total number of time-steps.\\
	\textbf{t}: Current time-step.	
	\begin{algorithmic}[1] 
		\STATE Calculated initial state $\rightarrow$  $s_{0}$.
		\FOR {$t \in [0, t_{max}]$ }
		\STATE Calculate action probabilities $\rightarrow$  $p(a|s_{t})$ 
		\STATE Make decision $\rightarrow$ $a_t$.
		\STATE Perform $a_t$.
		\STATE Calculated next state $\rightarrow$  $s_{t+1}$.		
		\STATE Calculate reward $\rightarrow$ $r_t$
		\STATE Calculate error $\rightarrow$ $TD_{\textrm{error}}$.
		\STATE Find closest component to $s_t$ $\rightarrow$ $m_{t}$. 
		\STATE Calculate similarity between $s_t$ and $m_{t}$ $\rightarrow$ $d_t$. 		
		\IF {$d_t$ $<$ $\rho_{t}$ OR $TD_{\textrm{error}}$ $>$ $T_{l}$ }
		\STATE Calculate weights $w$
		\STATE Update $m_t$
		\ELSE
		\STATE Create a new component.
		\ENDIF
		\STATE $s_{t}$ $\leftarrow$ $s_{t+1}$.
		\ENDFOR 
	\end{algorithmic}
\end{algorithm}

\subsubsection{Input description}
We base our input on the semantic segmented input image. The image of the segmentation map is separated into six different regions, for each we calculate a weighted histogram of the class distribution. The number of regions represents basic directional information (three vertical divisions) and distance information (two horizontal divisions) of the scene \cite{gharaee2016attention}. This approach can facilitate further studies of how agent learns to control its attention to each region and if an attention mechanism could improve the performance of the task \cite{Gharaee6, Gharaee7}. To further reduce the dimensionality of the input, the semantic labels are clustered into five categories as: \textit{Road}, \textit{Road-line}, \textit{Off-road}, \textit{Static object} and \textit{Dynamic object}.

The feature vector of each patch is concatenated into a single state-vector. The state-vector is normalized by its $\ell_1$-norm. Due to the low density of road lines in the semantic segmentation input image, we weighted the class, road lines, by 20 for the histogram calculations. 

In reality, ground-truth is obviously not available and as a result we need to estimate the semantic segmentation from available input, such as RGB. For our investigation, we utilize two different types of input data in our experimental setup, the ground truth semantic segmentation input directly from the simulator and the estimated semantic segmentation generated from RGB-images by EncNet \cite{HangZhang}. The EncNet model is trained offline on images collected from the CARLA simulator.

\textbf{EncNet }
We use an EncNet with the Resnet-101 as backbone architecture on top of which a special module named Context Encoding Module is stacked. The main reason behind the selection of EncNet over other powerful CNNs is because of the availability of pre-trained EncNet weights on the large and diverse ADE20K \cite{BoleiZhou}. In addition, EncNet has low computation complexity compared to CNNs such as PSPNet and DeepLabv3 and provides better inference speed at run time. 

\subsubsection{Decision making}
During training the agent applies an epsilon greedy policy to explore the world and to develop its learned concepts. Our decision making strategy is designed in a way to increase the exploration in the beginning of learning and to diminish as it progresses. Therefore the policy is gradually shifting from an epsilon greedy to a greedy one. When the learning is converged, the agent primarily exploits its learned concepts for decision making rather than exploring the world.

Our behavior policy is implemented in two steps. First, we use a greedy approach to select the greedy action $a_{gd} = \argmax_a \left( {p(a|s_{t}}) \right)$. Second, we sample an action from the distribution:

\begin{equation}
p_{\pi}(a|s_t)= \begin{cases}\vspace{8pt}
\frac{1-\tau}{|\mathcal{A}|}+\tau, & \text{if } a=a_{gd}\\ 
\frac{1-\tau}{|\mathcal{A}|}, & \text{else, }  \\
\end{cases}
\label{eq:wei}\\
\end{equation}

where $\tau \in [0, 1]$ is an increasing temperature to increase the probability of the greedy action as the learning progresses.

\subsubsection{Reward design}
We select four different reward signals representing important types of failures. These are Collision, Off-road, Opposite-lane and Low-speed. These failures generate the reward signal but only one of them is applied at a time based on its importance:
\begin{equation}
r = \begin{cases}
-r_{k_1}  & \text{if Collision} \\
-r_{k_2}\cdot r_o, & \text{else if Off-road } \\
-r_{k_3}\cdot r_l, & \text{else if Opposite-lane }\\
r_{\text{speed}}, & \text{else, }\\
\end{cases}
\label{eq:r1}
\end{equation}
where $r_o$ is calculated as the percentage of the car being off road and $r_l$ is the percentage of the car not being in the correct lane. The values of $r_o$ and $r_l$ are received from the CARLA simulator. Finally, $r_{\text{speed}}$ rewards the agent when it drives with a speed $v_t$ relative to the target speed $v_{\text{target}}$ at time step $t$:

\begin{equation}
r_\text{speed} = \begin{cases}
-r_{k_4} \cdot (\frac{v_t-v_{\text{target}}}{v_{\text{target}}})^2, & \text{if } v_t < 0\\
-r_{k_5} \cdot (\frac{v_t-v_{\text{target}}}{v_{\text{target}}})^2, & \text{if } 0<v_t < v_{\text{target}}\\
0, & \text{else. }\\
\end{cases}
\label{eq:rs}
\end{equation}

Additionally, a reward based on the road view, the percentage of the road visible in the image input to the agent, is always applied to align the agent with the road as: 

\begin{equation}
r_t = r+r_\text{road-view}.
\label{eq:r2}
\end{equation}

\subsubsection{Control signals}
The control signal that the simulator receives are, similarly to a real car, \textit{steering}, \textit{throttle}, \textit{brake} and a flag for the \textit{reverse} gear. Our actions are chosen to correspond to four action primitives that are able to fully control the vehicle velocity and direction, which are: \textit{Drive forward}, \textit{Turn to the right}, \textit{Turn to the left} and \textit{Drive backward}, each corresponds to a certain control signal.

\subsection{Experimental setup}
\label{sec:Expsu}

We train our system in a single simulated town, Town01 in CARLA. The training is done in the \textit{sunny noon} weather condition and it contains three different types of scenarios. The first scenario is driving along a road going straight forward and the other two are following the road through a right, respectively a left turn.
None of the scenarios include intersections or dynamic obstacles, e.g. other vehicles and pedestrians. The intersections are excluded from the training since our current system does not explicitly handle the multi-modality that arises from different possible choices of where to drive.

The CARLA simulator is able to simulate multiple sensor input types, RGB, depth and semantic segmentation images as well as Lidar point-clouds. For our system we are using either the semantic segmentation input directly or by estimating it from the RGB image using EncNet. 
The simulator also provides the measure necessary for calculating the reward signal, the exception being the road view that is based on the input state-vector.

The system is trained for a total number of time-steps, $t_{max}$, in order to allow it to converge. Each time-step is a single frame, but in order to get a reasonable state difference between time-steps only every seventh frame is used for training and decision making. The simulator itself is running at a frame rate of 7Hz.

We designed our experiments using three different scenarios, training, validation and test sets. The parameters of the system are set based on the training set and tuned on the validation set. The corresponding settings to run our experiments are presented by table \ref{tab:settings}. Based on the values shown in table \ref{tab:settings}, parameters, $\alpha$, $\tau$ and $\rho$ are being calculated at each time step: 

\begin{equation}
X_t \gets X_{Rate}(X_{Final}-X_t)+X_t, 
\label{eq:decay}
\end{equation}

where $X_t$ shows the values of the changing parameters, $\alpha$, $\tau$ and $\rho$, at the current time step $t$.

\begin {table}
\centering 
\begin{tabular}{ |p{1cm}|p{1cm}|p{1cm}|p{1cm}|p{1cm}|p{1cm}|   }
	\hline
	\multicolumn{6}{|c|}{Parameters Settings} \\
	\hline
	\hspace{0.5ex}
	$r_{k_1}$           & 50     & $\rho_{Init}$          & 0.1        & $\alpha_{Init}$ & 0.99\\
	\hspace{0.5ex}
	$r_{k_2}$           & 40     & $\rho_{Rate}$        & 3e-7      & $\alpha_{Rate}$ &1e-5\\
	\hspace{0.5ex}
	$r_{k_3}$           & 30     & $\rho_{Final}$        & 0.01       & $\alpha_{Final}$ &0.01\\
	\hspace{0.5ex}
	$r_{k_4}$           & 15     & $\tau_{Init}$          & 0.5         & $T_l$                & -10 \\
	\hspace{0.5ex}
	$r_{k_5}$           & 10     & $\tau_{Rate}$        & 7e-3       & $T_u$               & -5 \\
	\hspace{0.5ex}
	$\gamma$         & 0.9  & $\tau_{Final}$            & 0.99         &  $t_{max}$ & 4500     \\
	
	\hline
\end{tabular}
\caption {The table shows the settings of the parameters of BRL used for the experiments of this article:\\
	$t_{max}$: Total number of time steps to train a model.\\
	$\gamma$: Forgetting factor.\\
	$[T_{l}, T_{u}]$: Lower/upper thresholds for $TD_{\textrm{error}}$.\\
	$[r_{k_1}, \dots, r_{k_5}]$: Reward coefficients.\\
	$\alpha$: Learning rate.\\
	$\tau$: Exploration temperature.\\
	$\rho$: Similarity measure for updates.\\
	The parameters $\alpha$, $\tau$ and $\rho$ are initialized according to the initial value and updated according to (\ref{eq:decay}).}
\label{tab:settings}
\end{table}

\section{Results}

We use four different settings combining training and deployment, each contains nine models and we name them according to the following schedule:\\ 
\textbf{TGDG}: Training and Deployment w/ Ground-truth.\\ 
\textbf{TEDE}: Training and Deployment w/ Estimate.\\ 
\textbf{TGDE}: Training w/ Ground-truth, Deployment w/ Estimate.\\ 
\textbf{TEDG}: Training w/ Estimate, Deployment w/ Ground-truth.\\

We design the first set of experiments to test the basic functionalities of our system according to the methodology in the experimental setup section \ref{sec:Expsu}. In these experiments we also compare our system performance with conditional Imitation Learning (IL) \cite{codevilla2018end} and deep Reinforcement Learning (RL) \cite{Dosovitskiy17} by using the provided pre-trained models and evaluating them in our validation settings. The results of the corresponding experiments are presented in section \ref{testZGH}.

In the second set of experiments, we used test settings of the benchmark presented at CoRL 2017 \cite{Dosovitskiy17} in order to compare it to IL, RL and a Modular Pipeline (MP) \cite{Dosovitskiy17}.
We select the best performing model for each setup in the first set of experiments and use these for evaluation in the second set.
The results are presented in the section \ref{corl}.

It is important to mention that similarly to RL and IL our TEDE models are trained and deployed using the raw RGB input. Unfortunately, we cannot evaluate the potential benefit of training the compared models using semantic segmentation since the training codes of these methods are not provided and a comparison without training would not be fair. At the same time, the compared methods did only provide their best models, as such, we compare both to our best model as well as the average performance.

\begin{figure*}[h]
	\centering
	\begin{subfigure}[t]{0.28\linewidth}
		\centering
		\includegraphics[width=\linewidth]{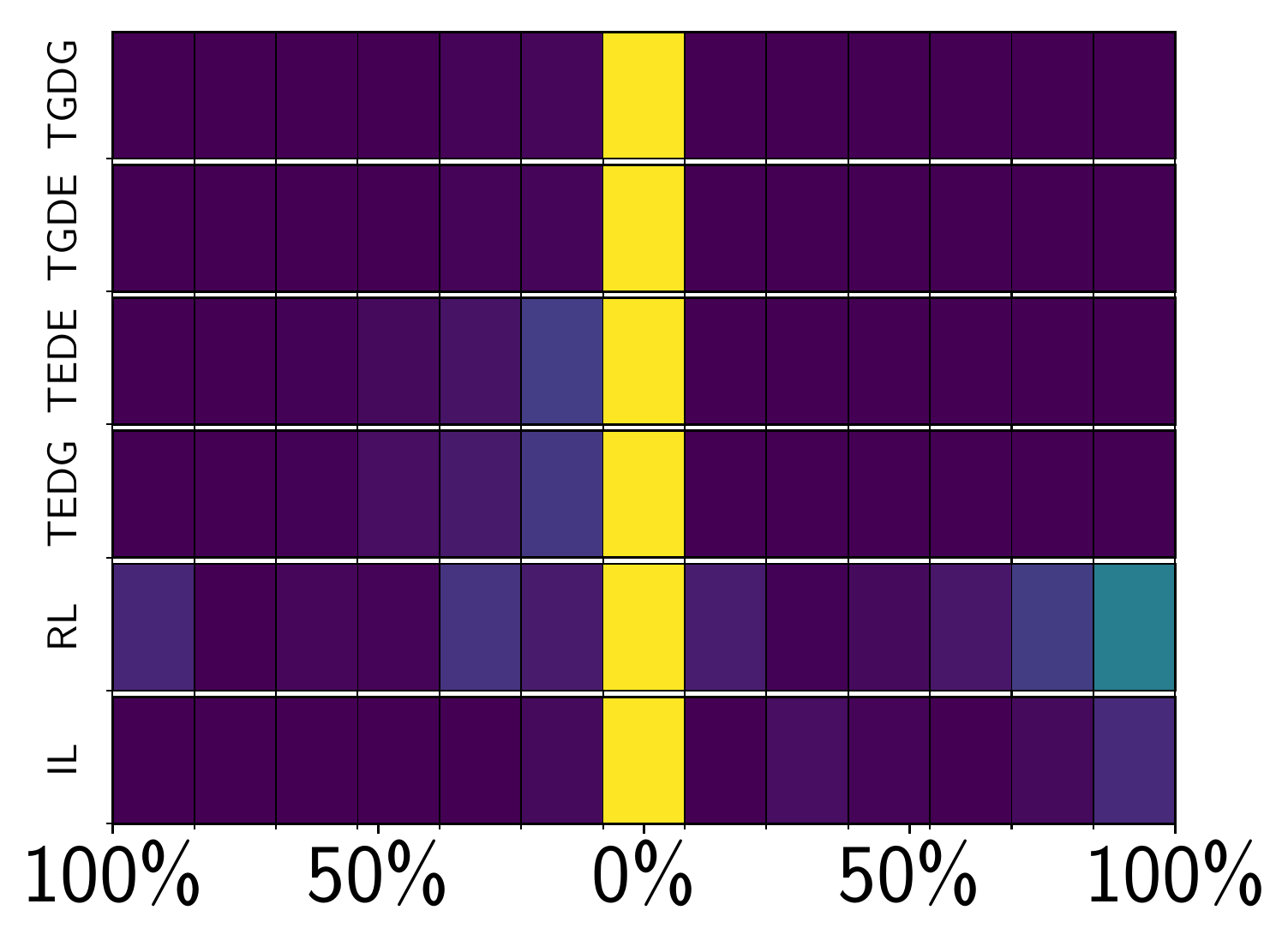}
		\caption{}
		\label{fig:roadHist}	
	\end{subfigure}
	\begin{subfigure}[t]{0.28\linewidth}
		\includegraphics[width=\textwidth]{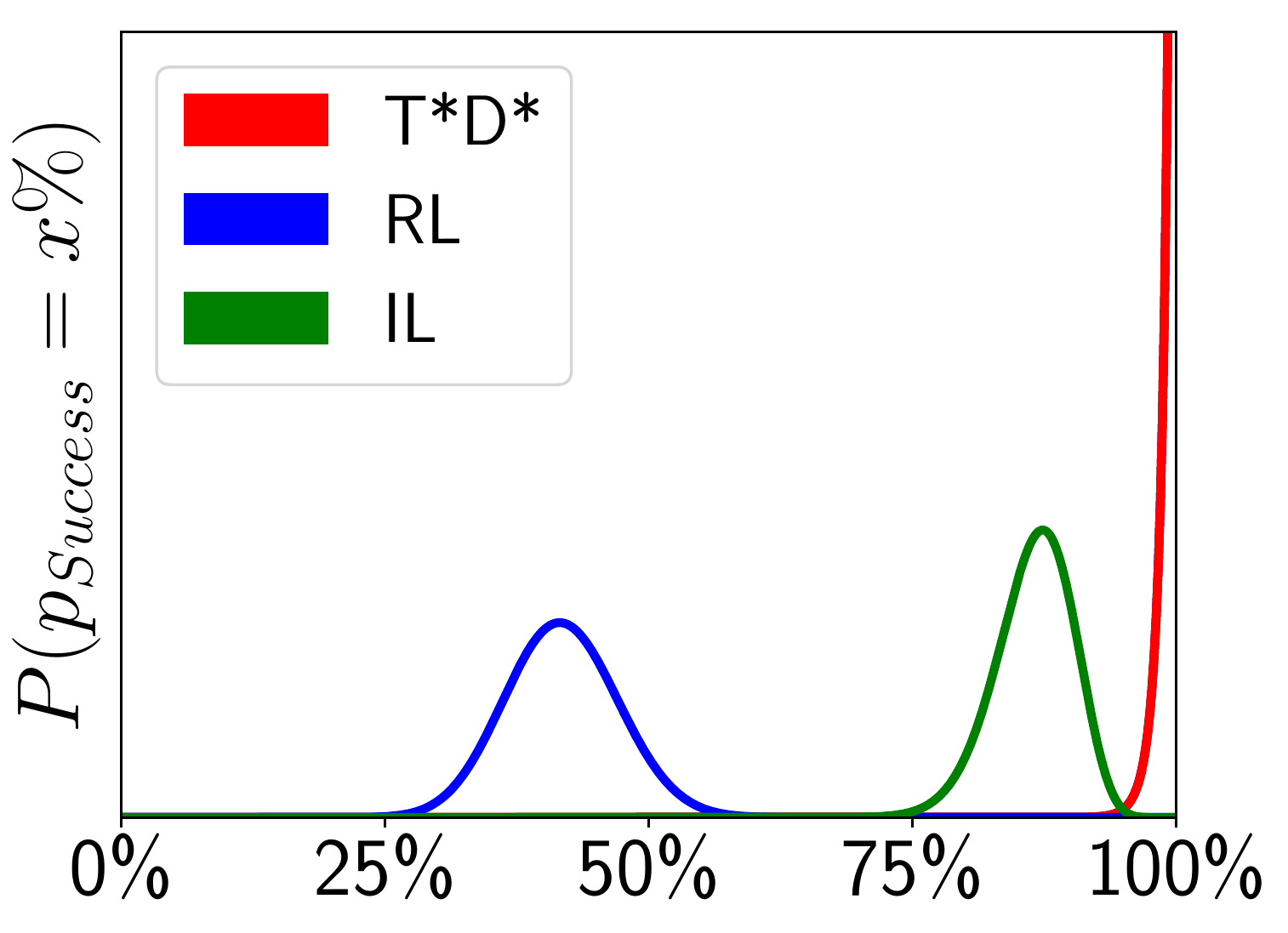}
		\caption{}
		\label{fig:BestSuccessBeta}
	\end{subfigure}
	\begin{subfigure}[t]{0.28\linewidth}
		\includegraphics[width=\textwidth]{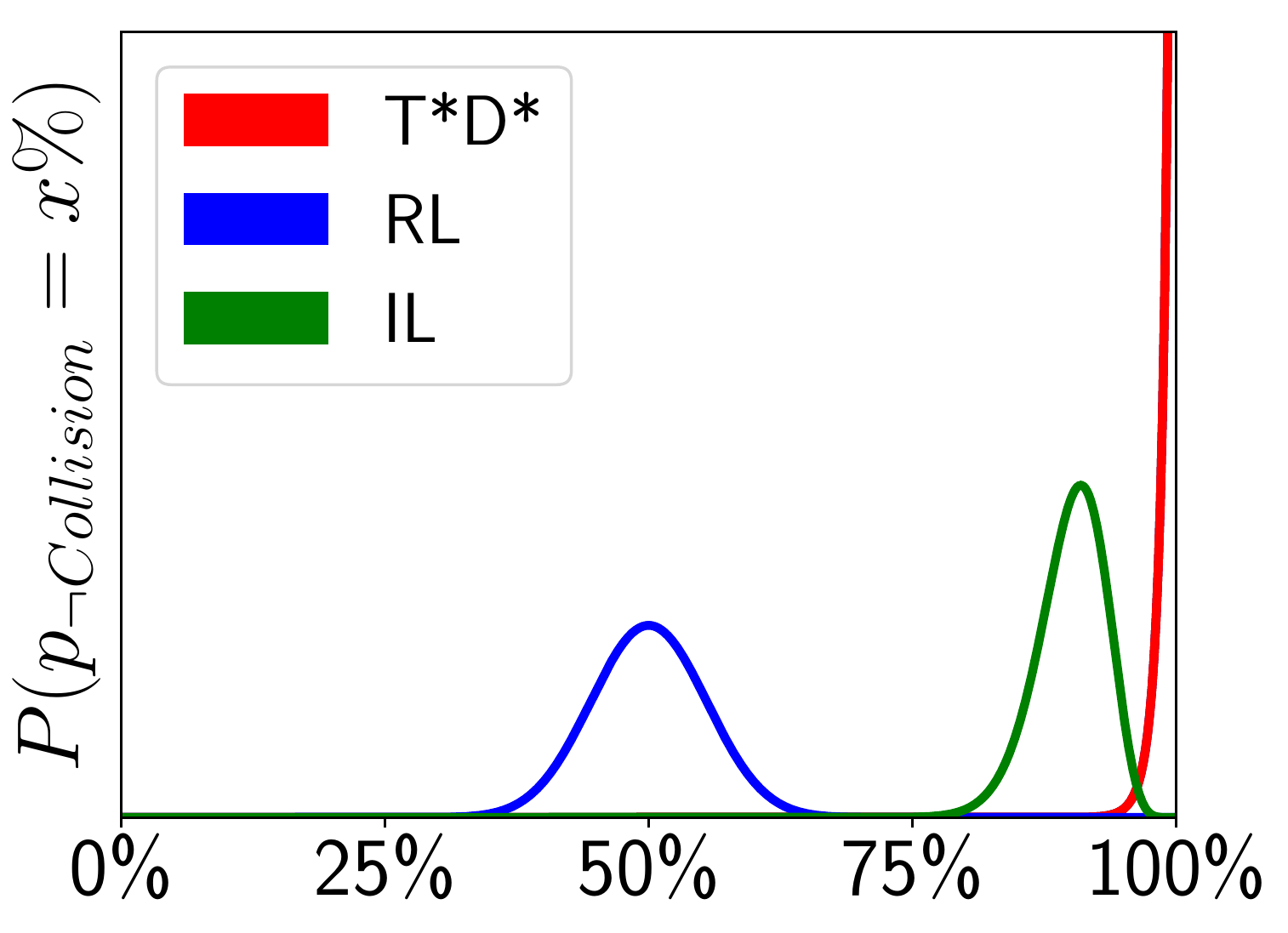}
		\caption{}	
		\label{fig:BestCollisionBeta}
	\end{subfigure}
	
	\begin{subfigure}[t]{0.28\linewidth}
		\centering
		\includegraphics[width=\linewidth]{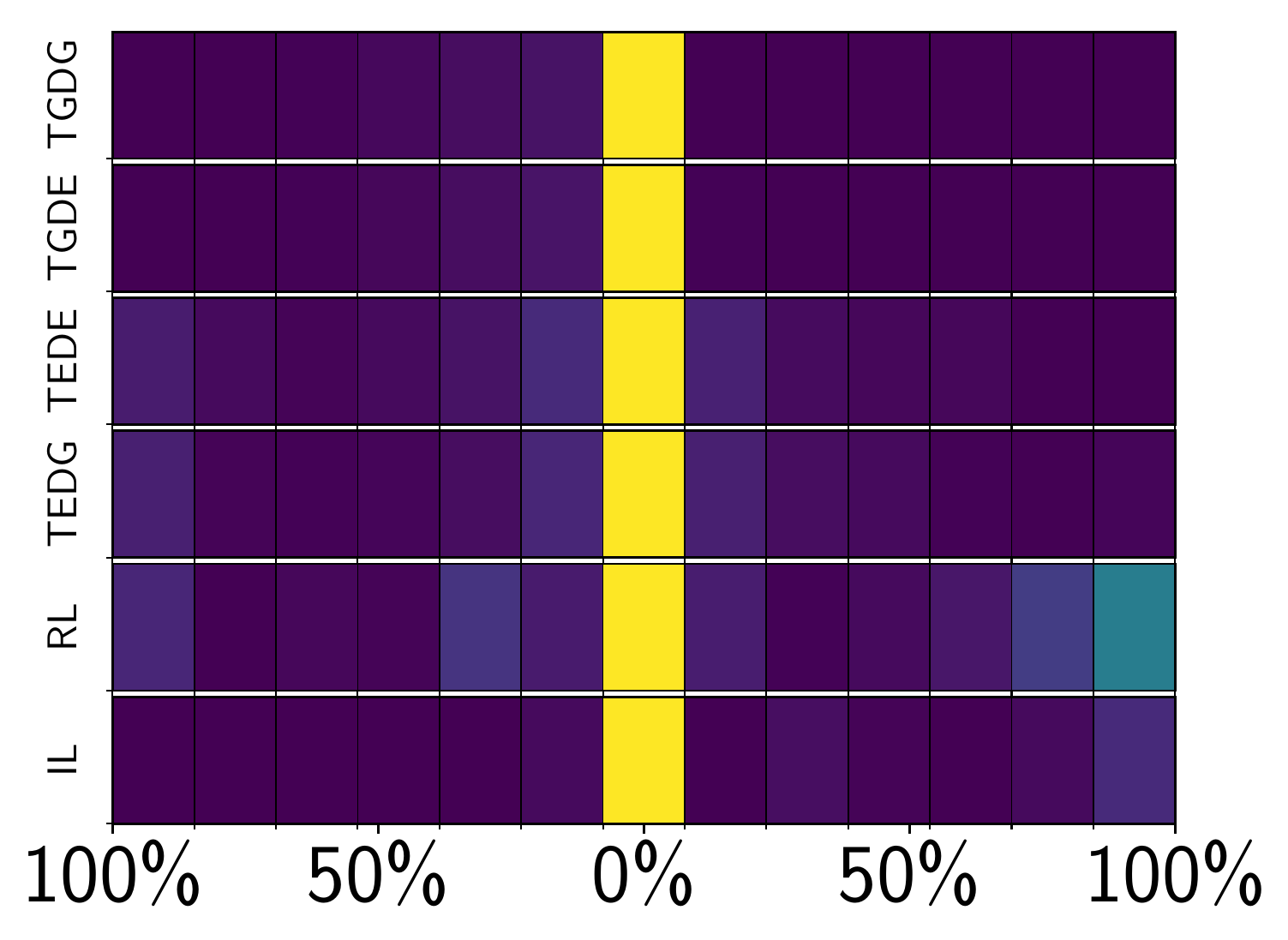}
		\caption{}
		\label{fig:roadHistAll}	
	\end{subfigure}
	\begin{subfigure}[t]{0.28\linewidth}
		\centering
		\includegraphics[width=\linewidth]{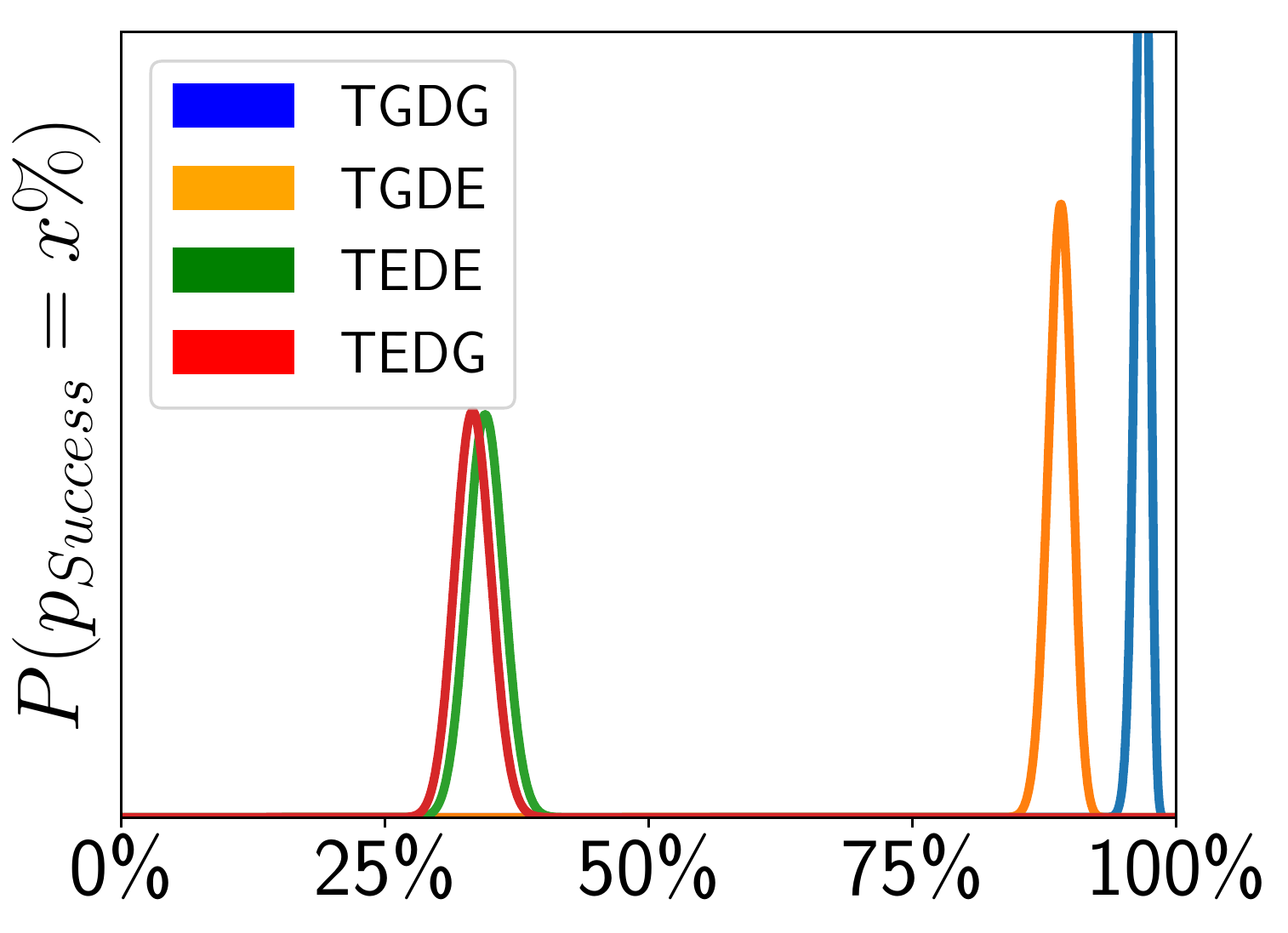}
		\caption{}	
	\end{subfigure}
	\begin{subfigure}[t]{0.28\linewidth}
		\includegraphics[width=\textwidth]{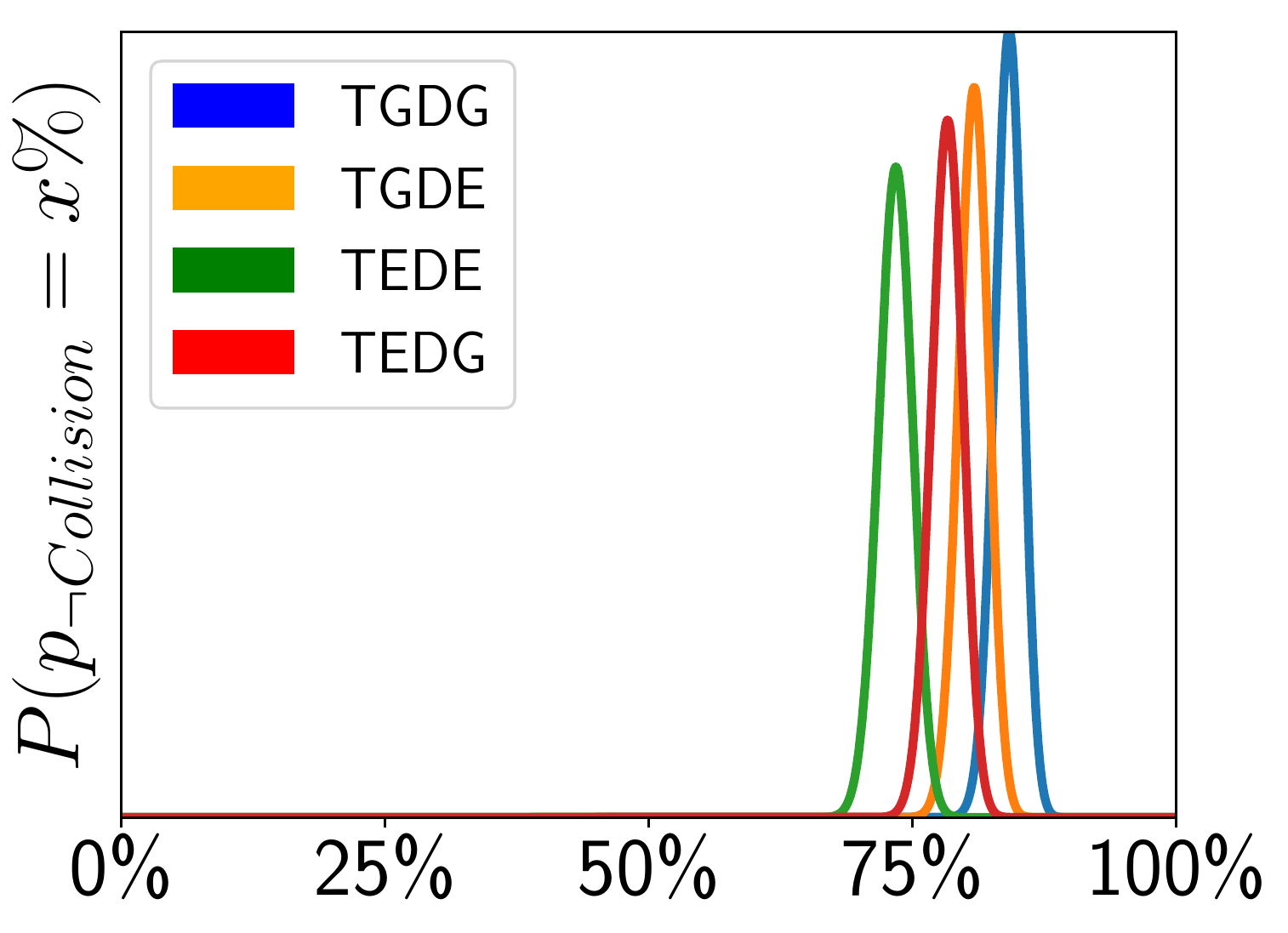}
		\caption{}	
	\end{subfigure}
	\caption{The results of the Best models are illustrated in the top row panels and the results from each of the model sets are illustrated in the bottom row panels. Histogram of the percentage of the car being other-lane/off-road shown in left column. Posterior probabilities of success rate in the middle column and collision avoidance in the right column are calculated using (\ref{eq:beta}). b-c) shows all four T*D*  models overlapping.}
	\label{fig:bestModels}
\end{figure*}

\begin{figure}[h!]
	\begin{subfigure}{0.84\linewidth}
		\includegraphics[width=\linewidth]{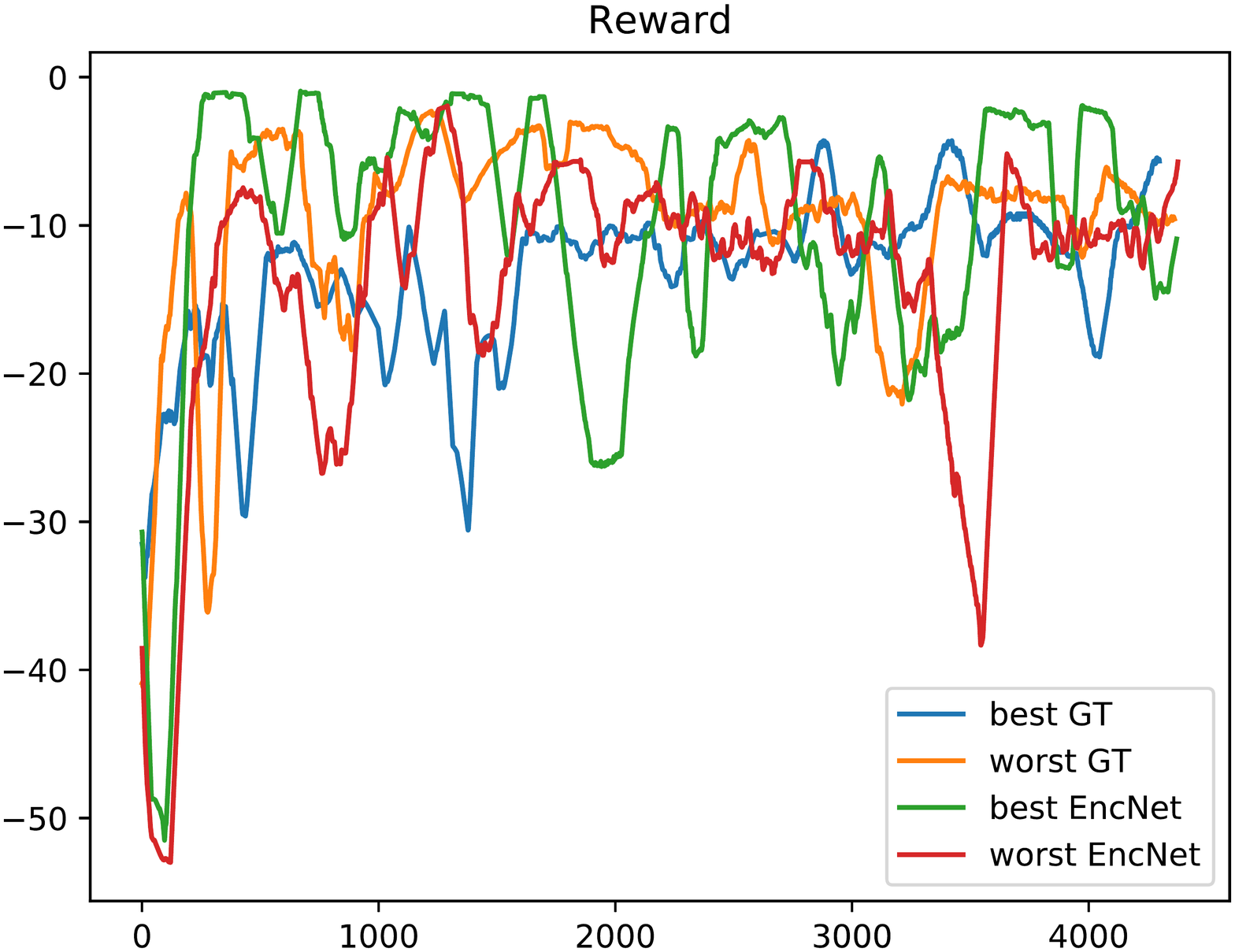}
		\caption{}
		\label{fig:reward}	
	\end{subfigure}
	\begin{subfigure}{0.84\linewidth}
		\includegraphics[width=\linewidth]{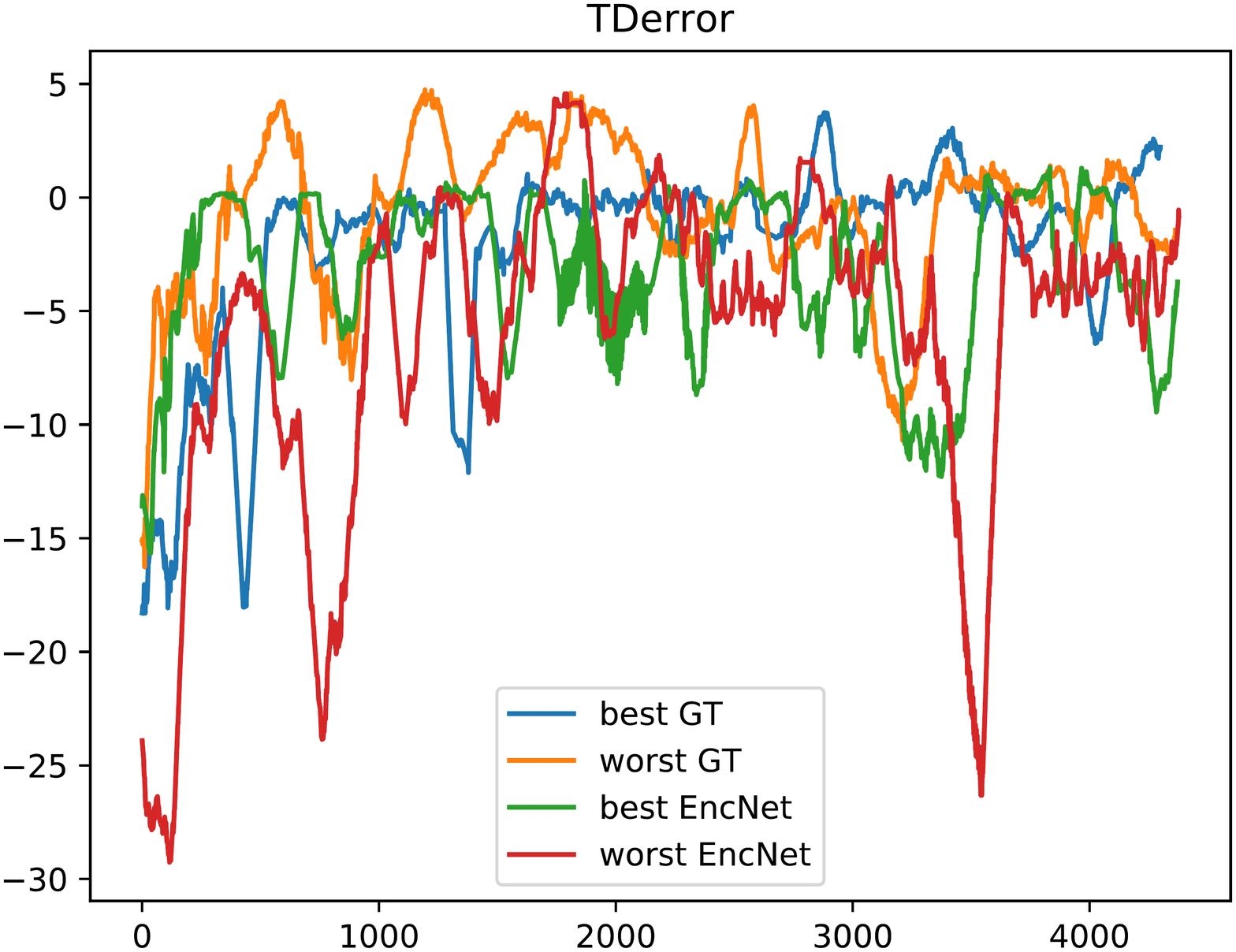}
		\caption{}
		\label{fig:tderr}	
	\end{subfigure}
	\caption{The reward signal convergence plots shown in (a) and the $TD_{err}$ signal convergence plots are shown in (b).}
	\label{fig:convres}
\end{figure}

Convergence plots showing the reward signal and the $TD_{err}$ for the best respectively the worst models trained with ground truth and estimated semantic segmentation are depicted in the figure \ref{fig:convres}.

\subsection{Test results}
\label{testZGH}

In this section, we present the results of the models evaluated according to the following metrics:

\begin{description}[leftmargin=!,labelwidth=\widthof{ No Collision  }]
	\item[Offroad] Being off-road (e.g, sidewalks). 
	\item[Otherlane] Being in the meeting lane.
	\item[Either]  Being off-road or in the meeting lane.
	\item[Success]  Accomplished tasks. 
	\item[No Collision]  Tasks without collisions.
	\item[Score] Average of \textit{Either}, \textit{Success} and \textit{No Collision}. 
	\item[Dist] Total distance driven in meter. 
\end{description}

A successful task is counted when the agent reaches its end within a certain amount of time. For each model set we calculate and report the average values of all metrics in Table \ref{tab:ZGHSmall}. We also determine the best performing models according to the \textit{Score} and report their performance.  We compare these results to the RL and IL methods for the same test setting.

\begin{table*}[h]
	\centering
	\begin{tabular}{c>{\rowmac}r>{\rowmac}c>{\rowmac}c>{\rowmac}c>{\rowmac}c>{\rowmac}c>{\rowmac}c>{\rowmac}c<{\clearrow}}
		\toprule
		Model & & Offroad & Otherlane & Either & Success & No collision & Score & Dist[m] \\
		\midrule
		\multirow{2}{*}{TGDG}
		& average  & 0.0\% & 11.1\% & 11.1\% & 96.8\% & 84.1\% & 0.90 ($\pm$0.07) &11132 ($\pm$703) \\
		& best model  & 0.0\% & 2.4\% & 2.4\% & 100.0\% & 100.0\% & 0.99 &11168 \\
		\midrule
		\multirow{2}{*}{TGDE}
		& average  & 0.1\% & 11.6\% & 11.7\% & 89.0\% & 80.8\% & 0.86 ($\pm$0.073) &11392 ($\pm$232) \\
		& best model  & 0.0\% & 2.5\% & 2.5\% & 100.0\% & 100.0\% & 0.99 &11119 \\
		\midrule
		\multirow{2}{*}{TEDE}
		& average  & 11.4\% & 29.8\% & 37.9\% & 34.5\% & 73.4\% & 0.57 ($\pm$0.3) &10209 ($\pm$1950) \\
		& best model  & 0.0\% & 17.5\% & 17.5\% & 100.0\% & 100.0\% & 0.94 &11227 \\
		\midrule
		\multirow{2}{*}{TEDG}
		& average  & 11.5\% & 25.8\% & 36.3\% & 33.3\% & 78.3\% & 0.58 ($\pm$0.3) &11437 ($\pm$2201) \\
		& best model  & 0.0\% & 22.7\% & 22.7\% & 100.0\% & 100.0\% & 0.92 &11238 \\
		\midrule
		\multirow{1}{*}{RL}
		&  & 42.4\% & 21.0\% & 52.4\% & 41.7\% & 50.0\% & 0.46 &8787 \\
		\midrule
		\multirow{1}{*}{IL}
		&  & 8.3\% & 0.7\% & 8.4\% & 86.9\% & 90.5\% & 0.90 &11293 \\
		\bottomrule
	\end{tabular}
	\vspace{0.5ex}
	\caption{Table shows the performance of the best model w.r.t score and allows a comparison between T*D*, IL and RL. The average is calculated per metric and shows the stability of the training for each of the T*D* settings. The values shown in the parenthesis for Score and Dist are the standard deviations of the models. Offroad, Otherlane and Either are calculated with a threshold of $20\%$ for each time-step.}
	\label{tab:ZGHSmall}
\end{table*}

Since classifying the vehicle as outside of the lane using a fixed threshold does not illustrate the gravity of the error we also illustrate the underlying distribution in figure \ref{fig:roadHist}. 
The left-part of the figure corresponds to other-lane and the right-part to off-road.

Figure \ref{fig:BestSuccessBeta} and \ref{fig:BestCollisionBeta} show the PDF (related to certainty) of our estimated success and collision rate using the beta distribution and Bayesian inference. As a uninformed prior, we select the Jeffrey prior of $\beta(0.5, 0.5)$ \cite{jeffreys1946invariant}, which puts a higher prior on either success, $s$, or failure, $f$, than a uniform $\beta(0,0)$ prior:

\begin{equation}
p(x|s,f) = \frac{x^{s-0.5}(1-x)^{f-0.5}}{\beta(s+0.5,f+0.5)}.
\label{eq:beta}
\end{equation}

In figure \ref{fig:bestModels}, bottom row, the different training and deployment models are evaluated. Instead of estimating the probability distribution parameters and calculating the histogram from the results of a single model the entire set of models is used.
The estimated distributions show the performance with respect to the set of models rather than a single model.

\subsection{Benchmark results}
\label{corl}

The benchmark proposed in \cite{Dosovitskiy17} is comprised by four different tasks, driving straight forward, one-turn (left or right), and two navigation tasks with multiple turns, each of them using the full road-network including intersections. All tasks except for the final navigation task are set in a static environment without vehicle and pedestrians while the last one contains multiple instances of each kind.
The reported metrics is the average kilometers driven between each type of infraction.

A comparison between our method and the three approaches presented in \cite{Dosovitskiy17} is shown in table \ref{tab:carlacompTown02}. Neither our nor the other methods have been trained on the environment in Town02 from the CARLA benchmark. 

\begin{table*}
	\centering
	\resizebox{\linewidth}{!}{%
		\begin{tabular}{l|ccccccc|ccccccc}  
			\toprule 
			& \multicolumn{7}{c}{New Town} & \multicolumn{7}{c}{New Town \& new weather} \\
			\midrule 
			Infraction type			& MP 	& IL 	& RL 	& TGDG 	& TEDE	& TEDG	& TGDE	
			& MP 	& IL 	& RL 	& TGDG	& TEDE 	& TEDG	& TGDE \\
			\midrule
			Opposite lane    		& 0.45 	& 1.12 	& 0.23	& 2.14 	& 0.18 	& 0.24 	& \textbf{4.10} 	
			& 0.40 	& 0.78 	& 0.21 	& 2.13 	& 0.18 	& 0.25 	& \textbf{2.93} \\ 
			Sidewalk    			& 0.46	& 0.76	& 0.43	& 0.40 	& \textbf{10.24} & 9.80 	& 0.11 	
			& 0.43	& 0.81 	& 0.48 	& 0.39	& 6.64 	& \textbf{9.80} 	& 0.10 \\ 
			Collision-static  		& 0.44	& 0.40  & 0.23	& \textbf{2.52} 	& 2.16 	& 2.18 	& 0.55 
			& 0.45	& 0.28	& 0.25 	& \textbf{3.55} 	& 1.53 	& 3.27 	& 0.79 \\
			Collision-vehicle	    & 0.51	& \textbf{0.59}	& 0.41	& 0.34 	& 0.22 	& 0.27 	& 0.38 	
			& \textbf{0.47}	& 0.44 	& 0.37 	& 0.33	& 0.26 	& 0.24 	& 0.34 \\
			Collision-pedestrian	& 1.40	& 1.88	& \textbf{2.55}	& 1.4 	& 1.20 	& 1.35 	& 0.53 	
			& 1.46	& 1.41 	& \textbf{2.99} 	& 1.39	& 0.69 	& 1.13 	& 1.31 \\
			\bottomrule
		\end{tabular}
	}
	\vspace{0.5ex}
	\caption{Performance evaluation in the (CoRL2017) Carla benchmark for Town02, the table shows average km driven between infractions per class (MP, IL and RL are evaluated in \protect\cite{Dosovitskiy17}). The total number of km might differ.}
	\label{tab:carlacompTown02}
\end{table*}

\section{Discussion}

According to the results, our system outperforms the compared methods on tasks without intersections. Our system also shows good performance on the benchmark settings.

\textbf{Test results}
According to table \ref{tab:ZGHSmall}, the best performing model out of nine, outperforms both RL and IL in terms of the scoring. We also see that the average score outperforms the RL-method.

Even though the average score of the TG* and TE* models differs, the results in figure \ref{fig:bestModels}, top row, show that the best model in each of the cases outperforms the best models of RL and IL. Figure \ref{fig:roadHist} also gives insight in how the lane positioning of the models compare. The RL model seems to either drive well inside of the lane or drive completely off-road while most of the other models stay close to the middle of the lane and slightly crosses over to the meeting lane.
Figures \ref{fig:BestCollisionBeta} and \ref{fig:BestSuccessBeta} show that the performance difference between our models is small while RL and IL has a much lower probability of success respectively not colliding.

The results in table \ref{tab:ZGHSmall} also clearly show that training on the ground truth semantic segmentation is beneficial even during deployment using the estimated semantic images (TGDE). TGDE significantly outperforms training and deploying using estimated input (TEDE). This result is somewhat surprising since much earlier work has indicated that using the same type of data for training and deployment is beneficial \cite{rusu2017sim, barrett2010transfer, saleh2018effective}. 
However, we further support these results in figure \ref{fig:bestModels}, bottom row, which shows a clear difference in success rate between TGDE- and TEDE-models. The figure also shows that the performance degradation of changing input is relatively small.

We hypothesize that the degradation of performance from changing input for training might be partly because of the over-segmentation of small details, which introduces a systematic noise in the feature extraction. If the system tries to over-fit to this noise it might struggle to learn the important concepts in the scene, and does lead to the inferior performance as observed in the experiments. This is also supported by the TDerror in figure \ref{fig:convres}, which shows that the TG models are more stable after convergance than the TE-models.

This result opens an interesting path in which a system could be trained from perfect, simulated data and then be applied outside of the simulator, possibly on real-world vehicles with minimal or no fine-tuning to adapt to the new environment.

\textbf{Benchmark}
Table \ref{tab:carlacompTown02} shows the results in a testing town in two distinct sets of weather conditions. For each column a higher average km between infractions is better and we see that our methods compare similarly to the other models.

The total training time for the system (based on 4500 time-steps) is in the order of 2-3h but using fewer time-steps or speeding up simulator clock are both possible approaches to reduce this time further. This can be compared to the training times mentioned in \cite{Dosovitskiy17}. The imitation learning approach is using 14h of recorded data while the reinforcement learning used roughly 12 days of non-stop driving at 10fps. The Modular pipeline is using a local planner and a PID controller both not requiring training. However, the perception component is based on a semantic segmentation model and trained using 2500 images from the CARLA simulator.

As shown in Table \ref{tab:carlacompTown02}, there are four infractions to evaluate the performance in the test settings. Among these infractions, opposite-lane and side-walk are highly anti-correlated, which means that a large improvement of the latter comes with a small degradation of the former. As a consequence we observe that in some cases, for example; side-walk, TE* models performs better than TG* models.

\section{Conclusion}

In this paper we presented a state-of-the-art performing reinforcement learning system for autonomous driving. The method is simultaneously clustering the agent's perceptual space based on its performed actions. The learning algorithm is developed based on a probabilistic Bayesian model, which enables the agent to deal with the uncertainty of its perception as well as the surrounding environment. Using a Gaussian Mixture Model with adaptive number of components, enables the agent to learn the action probabilities in the noisy environments. We have shown that by separating the input preprocessing learned offline from the online reinforcement learning method using semantic segmentation, the system efficiently learns the driving task.

The experiments results also show that by training our system using noise-free semantic segmentation input available from the ground-truth information in the CARLA simulator, we improve the training robustness as well as the test performance compared to estimating the semantic information from RGB-images. This result has also been shown to be the case when applying estimated input to the system trained with the ground-truth information.

\section*{Acknowledgment}
This work was partially supported by the SSF project RIT15-0097 and the
Wallenberg AI, Autonomous Systems and Software Program (WASP) funded by
the Knut and Alice Wallenberg Foundation.

\bibliographystyle{./IEEEtran} 

\balance

\bibliography{References}

\end{document}